\pdfoutput=1

\documentclass[11pt]{article}

\usepackage[final]{acl}

\usepackage{times}
\usepackage{latexsym}
\usepackage{tcolorbox}
\tcbuselibrary{skins, breakable, theorems}
\usepackage{booktabs}
\usepackage{multirow}
\usepackage{adjustbox}
\usepackage{xcolor}
\usepackage{soul}
\usepackage{graphicx}
\usepackage{subcaption}
\usepackage{caption}
\usepackage{makecell}

\usepackage[T1]{fontenc}

\usepackage[utf8]{inputenc}

\usepackage{microtype}

\usepackage{inconsolata}

\usepackage{graphicx}

\definecolor{MyBlue}{HTML}{a9d5ee}
\definecolor{MyLightBlue}{HTML}{DAEEFA}
\newcommand{\textcolorblue}[1]{
  \begingroup
  \sethlcolor{MyBlue}
  \textcolor{black}{\hl{#1}}
  \endgroup
}
\newcommand{\textcolorlblue}[1]{
  \begingroup
  \sethlcolor{MyLightBlue}
  \textcolor{black}{\hl{#1}}
  \endgroup
}

\title{Removal of Hallucination on Hallucination: Debate-Augmented RAG}

\author{
    \textbf{Wentao Hu}$^\spadesuit$\quad \textbf{Wengyu Zhang}$^\spadesuit$\quad \textbf{Yiyang Jiang}$^\spadesuit$ \\
    \textbf{Chen Jason Zhang}$^\spadesuit$\quad \textbf{Xiaoyong Wei}$^{\heartsuit, \spadesuit,}$\thanks{\ Corresponding author}\quad \textbf{Qing Li}$^\spadesuit$ \\
    $^\spadesuit$The Hong Kong Polytechnic University \quad $^\heartsuit$Sichuan University\\
    \texttt{wayne-wt.hu@connect.polyu.hk}\\
    \texttt{\{jason-c.zhang, cs007.wei\}@polyu.edu.hk}
}

\begin{document}
\maketitle

\begin{abstract}
Retrieval-Augmented Generation (RAG) enhances factual accuracy by integrating external knowledge, yet it introduces a critical issue: erroneous or biased retrieval can mislead generation, compounding hallucinations, a phenomenon we term \textit{Hallucination on Hallucination}.
To address this, we propose \textbf{D}ebate-Augmented \textbf{RAG} (\textbf{DRAG}), a training-free framework that integrates Multi-Agent Debate (MAD) mechanisms into both retrieval and generation stages.
In retrieval, DRAG employs structured debates among proponents, opponents, and judges to refine retrieval quality and ensure factual reliability.
In generation, DRAG introduces asymmetric information roles and adversarial debates, enhancing reasoning robustness and mitigating factual inconsistencies.
Evaluations across multiple tasks demonstrate that DRAG improves retrieval reliability, reduces RAG-induced hallucinations, and significantly enhances overall factual accuracy. Our code is available at \url{https://github.com/Huenao/Debate-Augmented-RAG}.

\end{abstract}
\section{Introduction}
Large Language Models (LLMs) have demonstrated remarkable natural language understanding and reasoning capabilities \cite{achiam2023gpt, touvron2023llama}.
However, their reliance on parametric knowledge introduces a critical challenge, hallucination, where the generated content deviates from factual correctness \cite{ji2023survey, huang2024survey}.
This issue severely limits their reliability, particularly in knowledge-intensive tasks where factual accuracy is paramount.

\begin{figure}[t]
    \centering
    \includegraphics[width=0.98\linewidth]{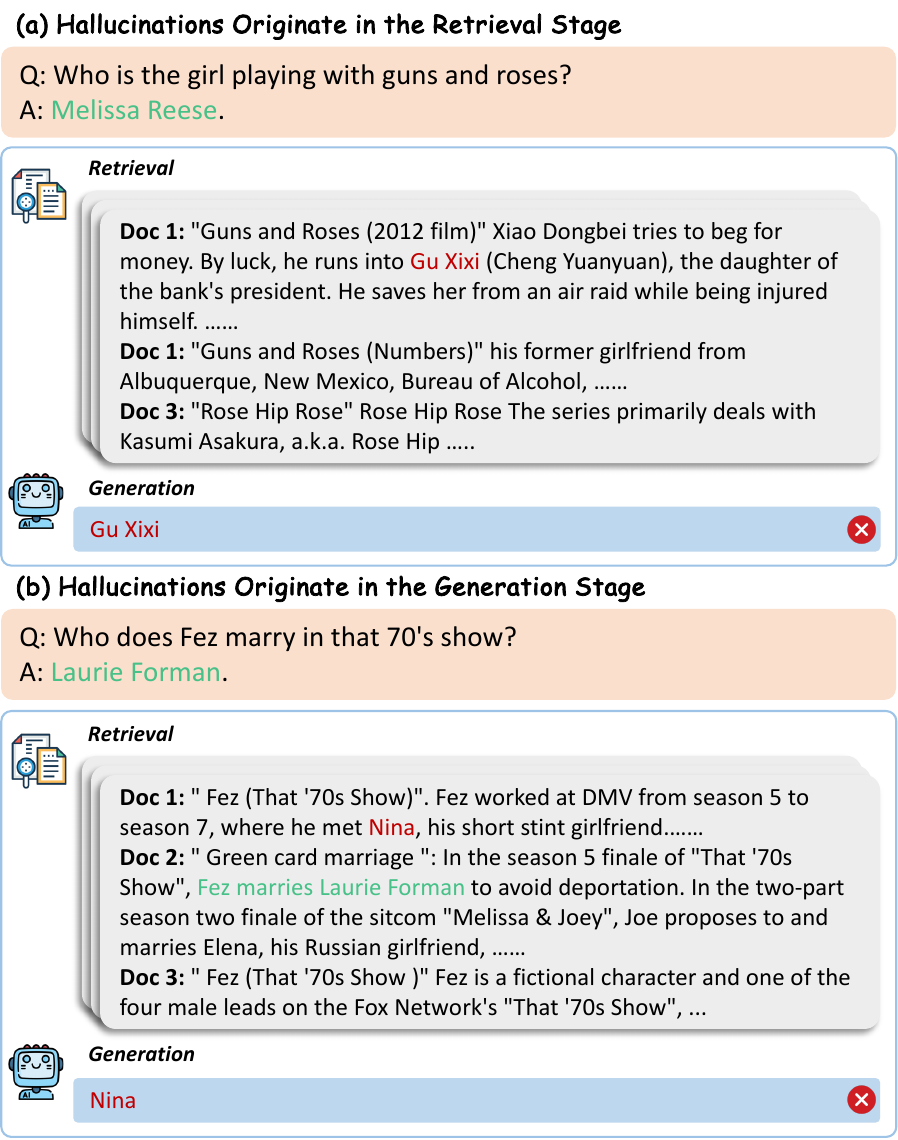}
    \caption{Demonstration of Hallucination in Retrieval-Augmented Generation. In the first example, the error stems from retrieving information about a movie with the same name instead of the correct entity (a band). In the second, despite retrieving accurate information, retrieval noise still leads to an incorrect response.}
    \label{fig:rag_hallucination_demo}
\end{figure}

To mitigate hallucinations, Retrieval-Augmented Generation (RAG) has been proposed as a framework that integrates external knowledge retrieval to enhance LLM outputs  \cite{gao2023retrieval}. 
By conditioning responses on retrieved documents, RAG reduces reliance on the model’s parametric knowledge, thereby aiming to improve factual correctness.
However, as shown in Figure~\ref{fig:rag_hallucination_demo}, RAG introduces a new challenge: \textit{Biased or erroneous retrieval results can mislead the generation, compounding the hallucination problem rather than solving it}.
In other words, the LLM’s inherent hallucinations are further amplified when retrieval provides incomplete, biased, or misleading information, leading to what we term \textit{Hallucination on Hallucination}. 
Addressing this challenge requires optimizing the entire RAG pipeline, encompassing both retrieval and generation stages.

In the retrieval stage, hallucinations frequently arise from insufficient retrieval, where incomplete or biased results mislead the generation phase. 
Even when retrieval strategies are optimized, the generation stage remains vulnerable to retrieval noise and misinformation \cite{ngo2007experimenting}, which can propagate hallucinations rather than mitigate them.
Existing approaches to reducing retrieval bias include iterative retrieval \cite{trivedi2023interleaving, shao2023enhancing, feng2024retrieval} and autonomous retrieval \cite{jiang-etal-2023-active, su-etal-2024-dragin, yu2024auto, wang-etal-2025-llms}, both of which seek to counteract hallucination stemming from knowledge deficiency and retrieval noise.
Additionally, reflection during generation \cite{asai2023self} and retrieval summarization \cite{kim2024sure} have been proposed to enhance the robustness and factual consistency of generation.
Existing methods do not systematically optimize the entire pipeline.
These methods also rely on LLMs for self-decision-making and self-optimization, making them susceptible to inherent biases, which limits their ability to generate diverse queries and robustness \cite{wei2008fusing}.

Recent advancements in Multi-Agent Debate (MAD) offer a promising paradigm for enhancing LLM robustness, factual accuracy, and reasoning diversity \cite{becker2024multi}.
MAD introduces multiple independent LLM agents that engage in structured debates, iteratively refining their responses through critical evaluation and multi-agent verification.
This process improves factual consistency and enhances robustness in reasoning \cite{du2023improving}.
Furthermore, MAD fosters perspective diversity by assigning distinct roles to agents, thereby mitigating biases that arise in single-agent approaches \cite{yin2023exchange, liang-etal-2024-encouraging}.

Motivated by this, we leverage MAD as a mechanism to address the challenge in RAG and propose \textbf{D}ebate Augmented \textbf{RAG} (\textbf{DRAG}).
DRAG is a novel training-free framework integrating multi-agent debate mechanisms across retrieval and generation phases. 
In the retrieval phase, multiple agents engage in structured debates to assess retrieval adequacy.
DRAG maintains a dynamic query pool by incorporating proponents advocating retrieval strategies, opponents questioning query sufficiency, and judges evaluating completeness, ensuring broader knowledge coverage and reduced factual biases.
In the generation phase, we establish asymmetric information roles among agents, reducing the LLM's over-reliance on retrieved content while promoting structured adversarial debate to enhance reasoning robustness and mitigate factual inconsistencies.
We evaluate DRAG on multiple tasks and demonstrate that this structured debate process leads to more reliable retrieval, enhanced reasoning robustness, and substantially reduced RAG-induced hallucinations.
Overall, our main contributions can be summarized as follows:
\begin{itemize}
    \item We propose a novel perspective on the challenges in RAG: \textit{Hallucination on Hallucination}.
    \item We introduce DRAG, a novel training-free framework that integrates Multi-Agent Debate (MAD) mechanisms into both retrieval and generation phases of RAG.
    \item We evaluate DRAG on multiple tasks, demonstrating its effectiveness in improving retrieval reliability, reasoning robustness, and reducing RAG-induced hallucinations.
\end{itemize}

\section{Related Work}
\subsection{Retrieval-Augmented Generation}
LLMs have demonstrated remarkable success across a wide range of tasks. 
However, they are prone to hallucinations, particularly when handling queries beyond their training data, such as in domain-specific or knowledge-intensive tasks.
To mitigate this issue, Retrieval-Augmented Generation (RAG) enhances LLMs by incorporating external knowledge retrieval into the response generation process \cite{gao2023retrieval}.
The most straightforward approach involves using the initial input as a query to retrieve information from an external corpus, which is then integrated into the model's input as supplementary knowledge \cite{guu2020retrieval, lewis2020retrieval, izacard-grave-2021-leveraging, izacard2022few, shi-etal-2024-replug}. 
However, single-round retrieval based solely on the initial input may fail to retrieve essential external knowledge for complex tasks and introduce harmful noise.

To meet complex retrieval needs, IRCoT \cite{trivedi2023interleaving} and ITER-RETGEN \cite{shao2023enhancing} introduce iterative retrieval-generation loops, allowing LLMs to refine their retrieval strategy dynamically to address insufficient retrieval.
Auto-RAG \cite{yu2024auto} further extends this by enabling autonomous query decision-making, enhancing retrieval flexibility.
FLARE \cite{jiang-etal-2023-active} and DRAGIN \cite{su-etal-2024-dragin} adopt a token-triggered retrieval mechanism, activating retrieval only when necessary and improving efficiency.
However, focusing on local tokens may neglect global reasoning needs. 
In contrast, Auto-RAG leverages LLM reasoning to optimize queries autonomously.
Self-RAG \cite{asai2023self} autonomously retrieves via reflection and follows a reflection-retrieval-regeneration process to improve factual consistency between generated and retrieved content.
RECOMP \cite{xu2023recomp} and SuRe \cite{kim2024sure} optimize LLM generation through summarized retrieval, enhancing the utilization of external information, reducing information overload, and improving factual consistency.
Existing methods lack a comprehensive analysis and optimization of RAG. Moreover, relying on a single-agent decision-making and reasoning process may introduce biases that impact performance.

\subsection{Multi-Agent Debate}
Recent studies have demonstrated that multi-agent interactions significantly enhance LLM capabilities across various dimensions \cite{becker2024multi}.
EoT \cite{yin2023exchange} mitigates biases and blind spots inherent in single-agent approaches by integrating external insights from other agents.
Moreover, Multi-Agent Debate (MAD) enables agents to iteratively challenge and refine each other’s claims, improving factual consistency \cite{du2023improving}.
Beyond factual alignment, MAD fosters divergent thinking, reducing reasoning stagnation and enhancing reasoning depth \cite{liang-etal-2024-encouraging}.
BDoG \cite{zheng2024picture} extends this paradigm to multimodal reasoning while introducing a blueprint to mitigate the trivialization of opinions and focus diversion.
The capabilities of MAD align well with the second-order hallucination problem in RAG, and this paper represents an initial endeavor to explore extending these capabilities to address this specific challenge.

\section{Debate-Augmented RAG}

\begin{figure*}
    \centering
    \includegraphics[width=0.98\linewidth]{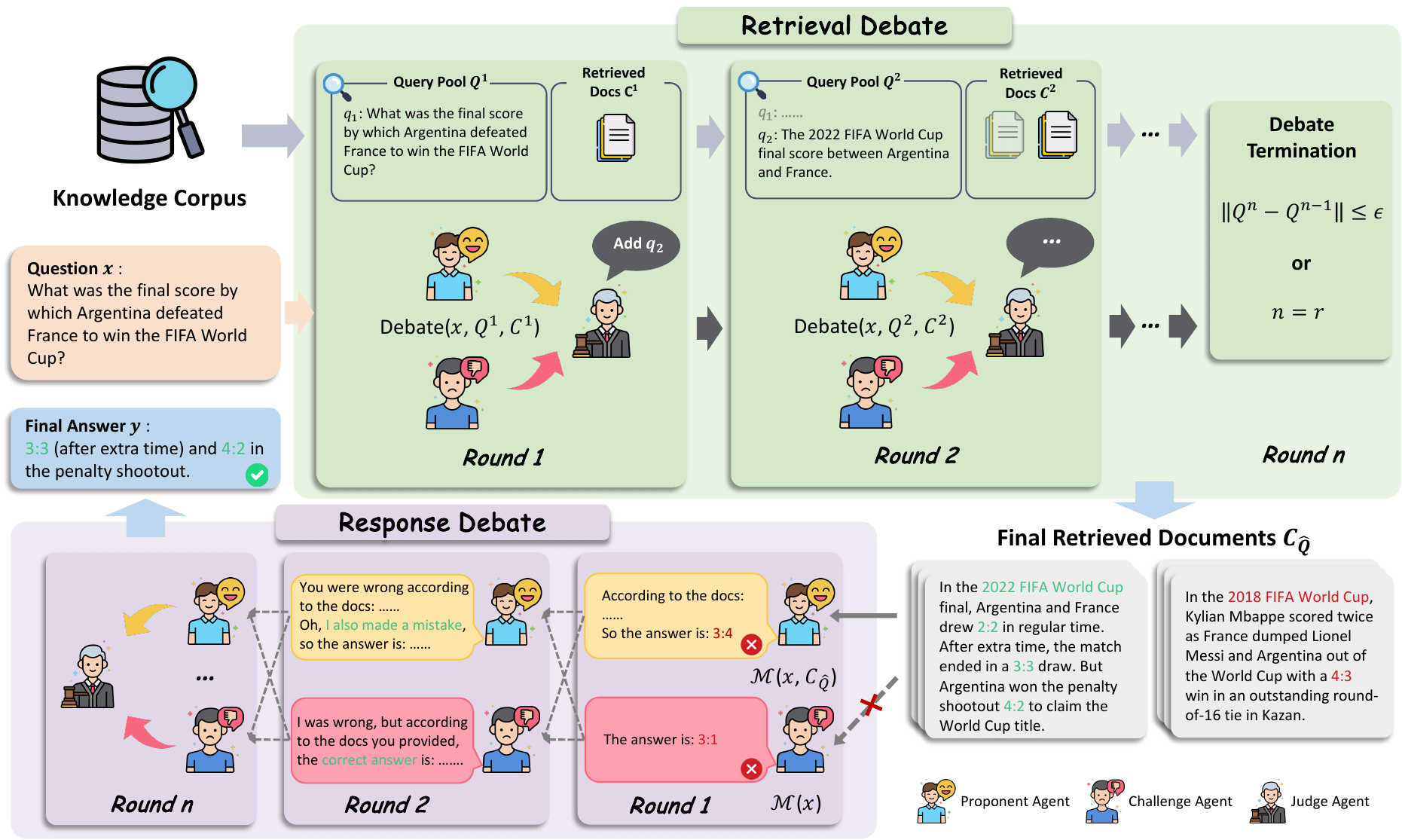}
    \caption{An overview of our Debate-Augmented RAG (DRAG) framework. It iteratively refines the retrieval strategy and enhances factual consistency.}
    \label{fig:framework}
\end{figure*}

In this section, we introduce Debate-Augmented RAG (DRAG), a general framework designed to enhance Retrieval-Augmented Generation (RAG) through a structured adversarial debate mechanism.

We first give an overview of DRAG in Section~\ref{sec:method_overview}, followed by a comprehensive discussion.
DRAG consists of a two-stage debate framework that comprehensively addresses the hallucinations introduced by RAG in both the retrieval and generation stages, as illustrated in Figure~\ref{fig:framework}.
Stage one is the Retrieval Debate, which is introduced in Section~\ref{sec:method_retdebate}, while stage two is the Response Debate, discussed in Section~\ref{sec:method_resdebate}.

\subsection{Overview}\label{sec:method_overview}
Formally, given a question $x$, a retriever $\mathcal{R}$ retrieves top-$k$ supporting paragraphs $C^{k}_{x}$ from a knowledge corpus $C$:
\begin{equation}
    \label{eq:1}
    C^{k}_{x} = \mathcal{R}(x, C, k)
\end{equation}
Once retrieval concludes, the retrieved paragraphs $C^{k}_{x}$ and the question $x$ are used to prompt a large language model (LLM) $\mathcal{M}$, which then generates an answer $y$:
\begin{equation}
    \label{eq:2}
    y = \mathcal{M}(x, C^{k}_{x})
\end{equation}

To mitigate hallucinations introduced at both retrieval and generation stages in RAG, DRAG employs a two-stage adversarial debate mechanism: \textbf{Retrieval Debate} and \textbf{Response Debate}. 
This debate-driven framework refines the retrieved content and ensures a robust, factually grounded response.
The following subsections introduce the debate mechanisms for each stage separately.

\subsection{Retrieval Debate}\label{sec:method_retdebate}
The retrieval process in standard RAG models often suffers from biases due to single-pass retrieval and suboptimal query formulation. 
DRAG introduces the Retrieval Debate $f_\text{RetDebate}$, a dynamic multi-agent debate mechanism to iteratively refine retrieval and retrieved contents: 
\begin{equation}
    \hat{Q}, C_{\hat{Q}} = f_\text{RetDebate}(x, \mathcal{R})
\end{equation}
where, $\hat{Q} = \{q_i\}^n_{i=1}$ represents the optimal query pool with $n$ queries for the question $x$, while $C_{\hat{Q}} = \{C^k_{q_i}\}^n_{i=1}$ denotes the corresponding retrieved paragraph set.

\noindent \textbf{Debate Structure.} The debate structure is a multi-round debate on optimizing the retrieval pool $Q$ between $m$ distinct agents $A = \{a_i\}^m_{i=1}$. The debate at the $j^{th}$ round can be formulated as follows:
\begin{equation}
    \mathcal{T}^j = {\tt Debate}(x, Q^j, C_{Q^j}, A)
\end{equation}
Each debate round aims to determine the appropriate refinement operation $\mathcal{T}^j$ applied to the retrieval pool $Q^j$, which is subsequently updated to $Q^{j+1}$ to initiate the next round.

\noindent \textbf{Agents and Roles.} Just like a real debate, this debate is also structured around agents having distinct roles assigned.
These different roles have distinct responsibilities, aiming to ensure a comprehensive and in-depth exploration of retrieval strategies by promoting perspective diversity and critical thinking, ultimately determining the optimal refinement operations for the retrieval pool.
In this stage, we define three agent roles.

\textbf{Proponent Agent} is required to provide justifications to support that the current query $Q^j$ is reasonable and that the retrieved results $C_{Q^j}$ are sufficient. Therefore, the operation supported by the proponent agent is:
\begin{equation}
\label{eq:5}
    \mathcal{T}_p:Q^{j+1} = Q^j
\end{equation}
that is, to keep $Q^j$ unchanged.

\textbf{Challenger Agent} critiques $Q^j$, identifying deficiencies and proposing modifications. Specifically, the challenger agent is responsible for proposing modifications to unreasonable queries or initiating new retrieval queries to address further knowledge needs.
So challenger agent insists on the refinement operation for the query pool $Q^j$:
\begin{equation}
\label{eq:6}
    \mathcal{T}_c : Q^{j+1} = (Q^j \cap Q_{\text{useful}}) \cup Q_{\text{new}} 
\end{equation}
where $Q_\text{useful} = \{q_i \in Q^j\}$ represents the subset of queries in $Q^j$ that the challenger agent deems necessary to retain. 
$Q_\text{new}$ consists of additional queries introduced by the challenger agent to supplement missing information or enhance retrieval quality.

\textbf{Judge Agent} evaluates the arguments presented by the proponent agent and the challenger agent and determines the ultimate operation $\mathcal{T}^j$ for the current round of debate:
\begin{equation}
    \mathcal{T}^j = {\tt Select}(\mathcal{T}_p, \mathcal{T}_c)
\end{equation}
where ${\tt Select}$ represents the decision-making function used by the judge agent. 

\noindent \textbf{Update Query Pool $Q$.} The debate starts with an initial query set $Q^0 = \{x\}$.
After each round of debate, the query pool $Q^j$ is iteratively refined following the judge agent’s decision and gets $Q^{j+1} = \mathcal{T}^j(Q^j)$, the specific operations as Eq.~\ref{eq:5} and Eq.~\ref{eq:6}.

The debate process continues iteratively until a convergence criterion is met:
\begin{itemize}
    \item The query set converges when $|| Q^{j+1} - Q^j || \leq \epsilon$, where $\epsilon$ is a predefined threshold ensuring minimal query modification.
    \item The maximum number of debate rounds $r$ is reached.
\end{itemize}

Once the debate terminates, the final query set $\hat{Q}$ is used for retrieval, forming the evidence set $C_{\hat{Q}}$ for the next stage, the Response Debate.

\subsection{Response Debate}\label{sec:method_resdebate}
Even after optimizing retrieval through the Retrieval Debate, the model may still be influenced by noisy or incorrect retrieval results.
To address this, DRAG introduces the Response Debate $f_\text{ResDebate}$, which iteratively enhances factual consistency and reasoning robustness:
\begin{equation}
\hat{y} = f_\text{ResDebate}(x, C_{\hat{Q}})
\end{equation}
where, $\hat{y}$ represents the final factually verified and robust response.

\noindent \textbf{Structure and Roles.}
As shown in Figure~\ref{fig:framework}, unlike the independent debates centered on the retrieval pool in the Retrieval Debate stage, the Response Debate fosters direct exchanges between agents.
To prevent over-reliance on retrieved information, the Response Debate employs an asymmetric information setting, ensuring agents engage in substantive argumentation rather than merely restating retrieved facts.
If all agents have equal access to retrieval results, they may uncritically accept the retrieved content, reinforcing biases instead of evaluating reliability.
Our experiments further confirm the effectiveness of this information asymmetry.

The information asymmetry is enforced through role assignments, defining three distinct agent roles:

\textbf{Proponent Agent} initiates the response based on retrieved information $C_{\hat{Q}}$ in first round:
\begin{equation}
    y^1_{A_p} = \mathcal{M}_{A_p}(x, C_{\hat{Q}} )
\end{equation}
and in the $i^{th}$ round, refines its response by incorporating the challenger agent's answer:
\begin{equation}
    y^i_{A_p} = \mathcal{M}_{A_p}(x, \{y^{<i}_{A_p}\}, y^{i-1}_{A_c})
\end{equation}
where $y^{i-1}_{A_c}$ represents the response from the challenger agent, which we defined later, in the $(i-1)^{th}$ round.

\textbf{Challenger Agent} generates an initial response based on internal knowledge:
\begin{equation}
    y^1_{A_c} = \mathcal{M}_{A_c}(x)
\end{equation}
and iteratively updates its response based on the proponent agent’s answer:
\begin{equation}
    y^i_{A_c} = \mathcal{M}_{A_c}(x, \{y^{<i}_{A_c}\}, y^{i-1}_{A_p})
\end{equation}

\textbf{Judge Agent}: Evaluates the final responses from both agents and selects the optimal answer:
\begin{equation}
    \hat{y} = \mathcal{M}_{A_j}(x, y^{r}_{A_p}, y^{r}_{A_c})
\end{equation}
where $r$ denotes the maximum number of debate rounds.

\section{Experimental Setup}
\subsection{Datasets}
We experimented on six benchmark datasets of three tasks. We follow previous work \cite{jiang-etal-2023-active} and randomly select 500 examples from each dataset for testing.

\noindent \textbf{Open-domain QA.} Open-domain QA aims to answer a question in the form of natural language based on large-scale unstructured documents. 
Here, we provide the model with only the questions without reference documents.
We choose three datasets for this task: NQ \cite{kwiatkowski2019natural}, TriviaQA \cite{joshi-etal-2017-triviaqa}, PopQA \cite{mallen-etal-2023-trust}. 
For the evaluation metrics, we extract the final answer from the output using regular expressions and calculate the exact match (EM) and token-level F1 score.

\noindent \textbf{Multi-hop QA.} Multi-hop QA focuses on answering questions that require the ability to gather information from multiple sources and conduct multi-step reasoning to arrive at a comprehensive answer.
We choose the 2WikiMultihopQA \cite{ho-etal-2020-constructing} and HotpotQA \cite{yang-etal-2018-hotpotqa}.
We also report EM and token-level F1 score for these datasets.

\noindent \textbf{Commonsense Reasoning.} Commonsense reasoning requires world and commonsense knowledge to generate answers.
We utilize the StrategyQA \cite{geva-etal-2021-aristotle} to evaluate DRAG and other baselines.
We extract the yes/no response and compare it with the standard correct answer using EM.

\subsection{Baselines}
We select \textbf{Naive Gen}, which is generated directly by an LLM, and \textbf{MAD} \cite{du2023improving} as baselines that do not incorporate retrieval.
For retrieval-based baselines, we use \textbf{Naive RAG}, which represents the standard RAG method without any optimization.
Additionally, we compare DRAG with RAG optimization baselines, including \textbf{IRCoT} \cite{trivedi2023interleaving}, \textbf{ITer-RetGen} \cite{shao2023enhancing}, and \textbf{FLARE} \cite{jiang-etal-2023-active}, which focus on improving the retrieval stage, as well as \textbf{Self-RAG} \cite{asai2023self} and \textbf{SuRe} \cite{kim2024sure}, which enhance the generation stage.

\begin{table*}[ht]
    \centering
    \adjustbox{max width=\textwidth}{
    \begin{tabular}{lcc|cc|cc|cc|cc|c}
        \toprule
        \multirow{2}{*}{Method} & \multicolumn{2}{c}{NQ} & \multicolumn{2}{c}{TriviaQA} & \multicolumn{2}{c}{PopQA} & \multicolumn{2}{c}{2Wiki} & \multicolumn{2}{c}{HotpotQA} & StrategyQA \\
        \cmidrule(lr){2-12}
        & EM & F1 & EM & F1 & EM & F1 & EM & F1 & EM & F1 & EM \\
        \midrule
        \multicolumn{12}{c}{\textit{Without Retrieval}} \\
        \midrule
        Naive Gen & 17.40 & 26.27 & 56.60 & 65.50 & 19.20 & 23.33 & 8.60 & 16.50 & 16.80 & 23.88 & 63.20 \\
        MAD & 21.80 & 33.11 & 56.40 & 66.39 & 21.40 & 28.28 & 18.20 & 25.13 & 23.00 & 32.79 &  \colorbox{MyLightBlue}{\textbf{66.20}} \\
        \midrule
        \multicolumn{12}{c}{\textit{With Retrieval}} \\
        \midrule
        Naive RAG & 38.20 & 50.08 & 60.80 & 69.55 & 37.60 & 45.69 & 14.80 & 24.27 & 25.80 & 35.80 & 62.60 \\
        IRCoT & 28.60 & 37.36 & 47.20 & 54.56 & 27.00 & 33.02 & \colorbox{MyLightBlue}{\textbf{22.80}} & \colorbox{MyLightBlue}{\textbf{31.19}} & 25.20 & 34.40 & 53.60 \\
        Iter-RetGen & \colorbox{MyLightBlue}{\textbf{40.80}} & \colorbox{MyBlue}{\textbf{52.31}} & \colorbox{MyBlue}{\textbf{63.00}} & \colorbox{MyBlue}{\textbf{72.23}} & \colorbox{MyLightBlue}{\textbf{39.60}} & 46.41 & 15.00 & 24.75 & \colorbox{MyLightBlue}{\textbf{27.80}} & \colorbox{MyLightBlue}{\textbf{38.93}} & 62.00 \\
        FLARE & 19.40 & 27.68 & 53.60 & 63.05 & 21.60 & 24.35 & 9.20 & 20.13 & 16.60 & 23.74 & 42.80 \\
        SuRe & 34.80 & 51.35 & 47.60 & 64.11 & \colorbox{MyBlue}{\textbf{41.80}} & \colorbox{MyBlue}{\textbf{48.99}} & 10.20 & 18.60 & 19.00 & 32.39 & 0.00 \\
        Self-RAG & \colorbox{MyBlue}{\textbf{44.00}} & \colorbox{MyLightBlue}{\textbf{52.20}} & 46.40 & 58.37 & 22.00 & 34.38 & 13.00 & 26.63 & 14.80 & 28.81 & 0.40 \\
        \midrule
        DRAG & 36.80 & 50.38 & \colorbox{MyLightBlue}{\textbf{60.80}} & \colorbox{MyLightBlue}{\textbf{69.93}} & 38.60 & \colorbox{MyLightBlue}{\textbf{46.50}} & \colorbox{MyBlue}{\textbf{28.80}} & \colorbox{MyBlue}{\textbf{36.97}} & \colorbox{MyBlue}{\textbf{30.80}} & \colorbox{MyBlue}{\textbf{41.74}} & \colorbox{MyBlue}{\textbf{69.20}} \\
        \bottomrule
    \end{tabular}
    }
    \caption{The overall evaluation results of DRAG and other baselines on six benchmarks. \textcolorblue{Blue} marks the \textcolorblue{best-performing} method, \textcolorlblue{light blue} represents the \textcolorlblue{second-best-performing} method. All methods are evaluated under the same settings.}
    \label{tab:main}
\end{table*}

\subsection{Implementation Details}
We selected Llama-3.1-8B-Instruct \cite{dubey2024llama} as our base LLM model.
Our implementation of DRAG and the baselines are built upon FlashRAG \cite{FlashRAG}, a Python toolkit designed for the reproduction and development of RAG-based systems.
Following FlashRAG, we employed E5-base-v2 \cite{wang2022text} as the retriever and used the widely adopted Wikipedia dump from December 2018 as the retrieval corpus \cite{karpukhin-etal-2020-dense}.
We retrieve top-$3$ paragraphs for each query for our DRAG and other iterative baselines.
The maximum debate interactions $r$ is set to $3$.
We set the number of agents per stage to $3$, with one agent assigned to each role.
The threshold $\epsilon$ is set to $0$.

\section{Experimental Results}

\subsection{Comparison with Baselines}
We conduct comprehensive experiments to evaluate the performance of DRAG against various baselines across six benchmark datasets. 
As summarized in Table~\ref{tab:main}, DRAG achieves strong and consistent improvements on multi-hop reasoning tasks, underscoring its robustness in settings that demand complex retrieval and multi-step inference. In comparison, on single-hop QA tasks, DRAG exhibits competitive performance, ranking second on several benchmarks with only marginal differences from the top-performing methods.

On multi-hop reasoning tasks such as 2WikiMultihopQA and HotpotQA, DRAG achieves substantial improvements in Exact Match (EM) scores, outperforming the best baseline methods by $6$ and $3$, respectively.
This superior performance can be attributed to DRAG’s structured multi-agent debate mechanism, which enhances retrieval adequacy and improves generation robustness.
Compared to generation-focused models such as SuRe and Self-RAG, DRAG demonstrates that merely optimizing the generation stage is insufficient for complex reasoning tasks; instead, retrieval quality plays a crucial role.
This aligns with our hypothesis that low-quality and insufficient queries negatively impact RAG model performance.
Additionally, when compared to retrieval-optimized models such as IRCoT and Iter-RetGen, DRAG exhibits a stronger ability to dynamically refine retrieval strategies, ensuring broader and more relevant knowledge retrieval.

While DRAG excels in multi-hop reasoning tasks, its performance on single-hop tasks is slightly lower than some retrieval-optimized baselines.
We attribute this to the phenomenon of problem drift induced by the debate mechanism \cite{becker2024multi}.
Specifically, excessive debates among agents may introduce unnecessary complexity, generating additional queries or conflicting perspectives that deviate from the optimal answer.
However, this issue can be effectively mitigated by adjusting the number of debate rounds, as further analyzed in Section~\ref{sec:an_rounds}.

\subsection{Analysis of Debate Rounds}\label{sec:an_rounds}

\begin{table}[t]
    \centering
    \adjustbox{max width=\linewidth}{
     \begin{tabular}{lc|c|c|c|c}
        \toprule
        \multirow{2}{*}{Settings} & NQ & TriviaQA & PopQA & 2Wiki & HotpotQA \\
        \cmidrule(lr){2-6}
        & EM & EM & EM & EM & EM \\
        \midrule
        DRAG$_{\text{RetIter=0}}$& 34.40 & 60.40 & \textbf{37.60} & 26.00 & 30.20 \\
        DRAG$_{\text{RetIter=1}}$& 36.80 & \textbf{60.80} & 37.40 & 29.20 & \textbf{31.40} \\
        DRAG$_{\text{RetIter=2}}$& 36.80 & 60.76 & 37.50 & \textbf{29.60} & 30.80 \\
        DRAG$_{\text{RetIter=3}}$& 36.80 & \textbf{60.80} & \textbf{37.60} & 28.80 & 30.80 \\
        DRAG$_{\text{RetIter=4}}$& \textbf{37.00} & 60.80 & \textbf{37.60} & 29.00 & 31.00 \\
        \midrule
        DRAG$_{\text{ResIter=0}}$& \textbf{38.20} & 57.20 & \textbf{39.60} & 16.00 & 28.20 \\
        DRAG$_{\text{ResIter=1}}$& 37.60 & \textbf{62.40} & 39.20 & 27.80 & \textbf{33.40} \\
        DRAG$_{\text{ResIter=2}}$& \textbf{38.20} & 60.20 & 38.80 & 28.20 & 31.20 \\
        DRAG$_{\text{ResIter=3}}$& 36.80 & 60.80 & 38.60 & \textbf{28.80} & 30.80 \\
        DRAG$_{\text{ResIter=4}}$& 37.00 & 60.60 & 38.80 & 27.20 & 31.00 \\
        \bottomrule
    \end{tabular}
    }
    \caption{Model performance with respect to the iteration round of debate.}
    \label{tab:eval_rounds}
\end{table}

To systematically assess the impact of debate rounds on DRAG's performance, we analyze the effects of Retrieval Debate and Response Debate rounds separately.
In each experiment, we varied the number of rounds in one stage (either retrieval or response) from 0 to 4, while keeping the number of rounds in the other stage fixed at 3.

\noindent \textbf{Retrieval debates help refine the retrieval pool.}
Table~\ref{tab:eval_rounds} shows that increasing the Retrieval Debate rounds generally enhances performance by progressively refining the retrieval pool.
For instance, on 2WikiMultihopQA, performance improves from 26.00 (RetIter=0) to 29.60 (RetIter=2) before stabilizing.
This trend indicates that iterative refinement helps mitigate retrieval noise and improves knowledge coverage.
However, beyond a certain threshold (e.g., RetIter=3), additional debate rounds contribute marginal gains. 
We attribute this to the convergence of the retrieval pool in the Retrieval Debate stage.

\noindent \textbf{Response debates enhance reasoning but risk problem drift.}
In the Response Debate stage, performance initially improves as adversarial discussions enhance factual consistency, but excessive debate rounds lead to performance degradation.
This is particularly evident in TriviaQA, where the EM score drops from 62.4 (ResIter=1) to 60.20 (ResIter=2), suggesting that early problem drift can emerge when excessive rounds introduce unnecessary complexity.
Similar effects are observed in HotpotQA and 2WikiMultihopQA, reinforcing the idea that over-iterating debates may amplify inconsistencies rather than resolve them.


\begin{table}[t]
    \centering
    \adjustbox{max width=\linewidth}{
    \begin{tabular}{c|l|cc}
    \toprule
     & Benchmark & \makecell{Debate Rounds} & \makecell{Query Counts} \\
    \midrule
    \multirow{4}{*}{\textit{Single-hop}} & NQ & 1.27 & 1.06 \\
    & TriviaQA & 1.33 & 1.08 \\
    & PopQA & 1.31 & 1.07 \\
    & StrategyQA & 1.18 & 1.07 \\
    \midrule
    \multirow{2}{*}{\textit{Multi-hop}} & 2Wiki & \textbf{1.57} & \textbf{1.17} \\
    & HotpotQA & \textbf{1.46} & \textbf{1.17} \\
    \bottomrule
    \end{tabular}
    }
    \caption{The average number of debate rounds and the counts of queries in the retrieval debate stage of DRAG on single-hop QA tasks and multi-hop QA tasks.}
    \label{tab:analysis_retdebate}
\end{table}


\subsection{Analysis of the Retrieval Debate}\label{sec:an_retdebate}

\noindent \textbf{DRAG dynamically adapts retrieval strategies to task complexity.}
To examine DRAG’s retrieval debates, we analyze debate rounds and query counts across single-hop and multi-hop QA tasks. Multi-hop QA requires retrieving interrelated evidence, demanding more iterative refinement than single-hop QA, which typically retrieves a single fact.
Table~\ref{tab:analysis_retdebate} shows a clear adaptation pattern in DRAG’s retrieval behavior.
Table~\ref{tab:analysis_retdebate} shows that for single-hop QA, DRAG converges quickly with minimal refinement (1.18 to 1.33 debate rounds, 1.06 to 1.08 queries per instance). For multi-hop QA, it engages in more debates (1.57 rounds for 2Wiki, 1.46 for HotpotQA) and higher query counts (1.17 rounds), ensuring adequate contextual retrieval.
These results highlight DRAG’s adaptive retrieval mechanism, dynamically adjusting debate intensity to optimize efficiency and retrieval sufficiency based on task complexity.

\begin{table}[t]
    \centering
    \adjustbox{max width=\linewidth}{
    \begin{tabular}{lccccc}
        \toprule
        & NQ & TriviaQA & 2Wiki & HotpotQA \\
        \midrule
        Question Counts & 139 & 176 & 338 & 331 \\
        \midrule
        \textbf{w/o $f_\text{ResDebate}$} & & & & \\
        Avg EM & 3.60 & 12.50 & 7.10 & 9.32 \\
        Avg F1 & 12.93 & 23.42 & 18.01 & 19.04 \\
        \midrule
        \textbf{with $f_\text{ResDebate}$} & & & & \\
        Avg EM & \textbf{5.04} & \textbf{28.98} & \textbf{19.23} & \textbf{17.36}  \\
        Avg F1 & \textbf{18.50} & \textbf{40.21} & \textbf{28.52} & \textbf{29.64} \\
        \bottomrule
    \end{tabular}
    }
    \caption{Comparison of performance with and without $f_\text{ResDebate}$ across different datasets.}
    \label{tab:comp_resdebate}
\end{table}

\subsection{Analysis of the Response Debate}
\noindent \textbf{Response debate enhances robustness against retrieval deficiency.}
To further examine DRAG’s ability to handle incomplete retrieval, we calculate the average EM with and without the Response Debate on a subset of questions where the gold answer is absent from the retrieved documents.
As shown in Table~\ref{tab:comp_resdebate}, this result further reinforces DRAG’s advantage over conventional RAG models by mitigating the impact of incomplete retrieval and ensuring a more robust and verifiable response generation process.

\begin{figure*}[t]
    \centering
    \includegraphics[width=0.98\linewidth]{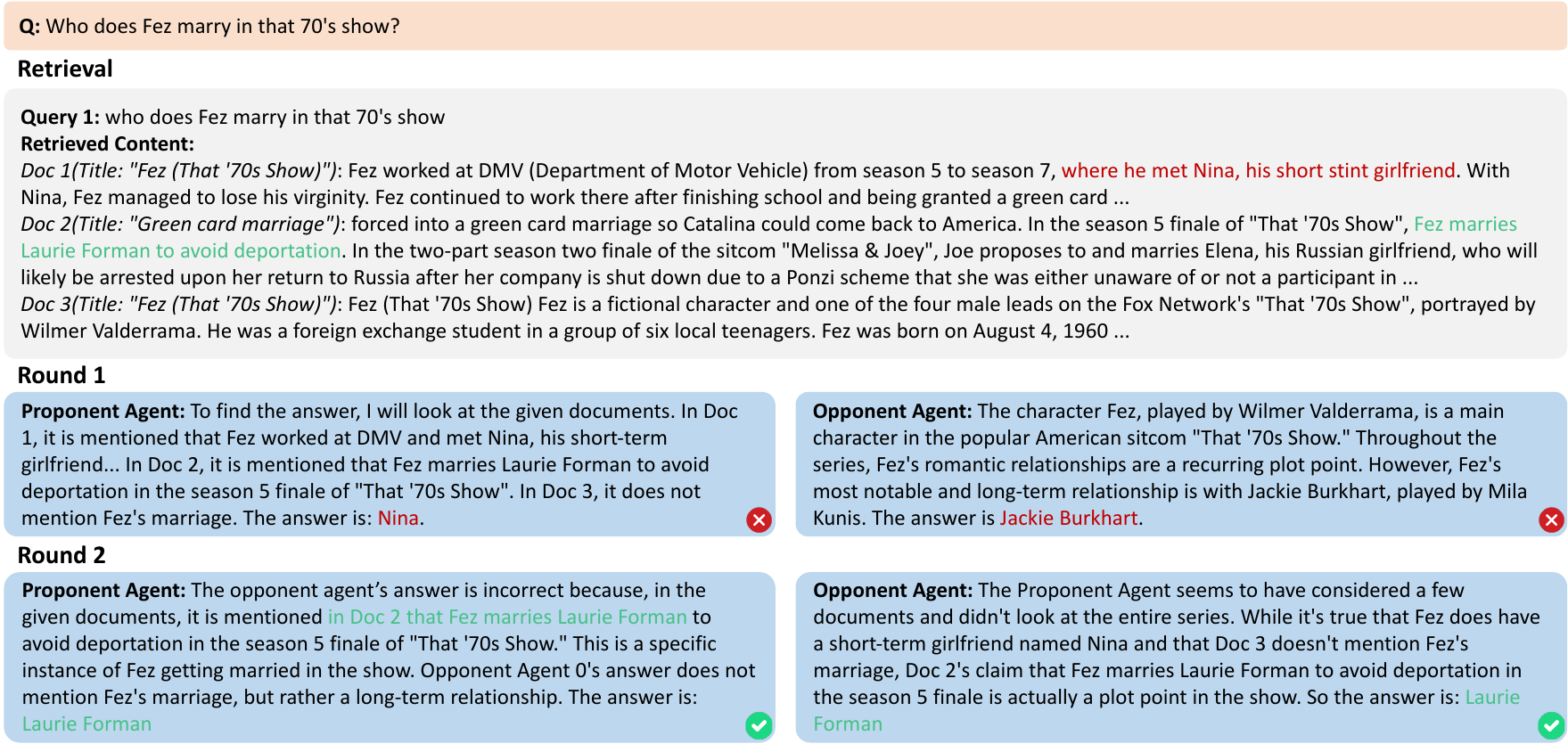}
    \caption{Case study of the response debate. }
    \label{fig:case_study1}
\end{figure*}

\noindent \textbf{Enhances the final response's alignment with the retrieved evidence.}
To assess the impact of response debates on factual consistency, we analyze the Response Debate stage in DRAG.
As shown in Figure~\ref{fig:case_study1}, this case study illustrates how multi-agent debates iteratively refine responses, correcting inaccuracies and reinforcing evidence-based reasoning.

\begin{table*}[t]
    \centering
    \adjustbox{max width=\textwidth}{
    \begin{tabular}{lcc|cc|cc|cc|cc|c}
        \toprule
        \multirow{2}{*}{Method} & \multicolumn{2}{c}{NQ} & \multicolumn{2}{c}{TriviaQA} & \multicolumn{2}{c}{PopQA} & \multicolumn{2}{c}{2wiki} & \multicolumn{2}{c}{HotpotQA} & StrategyQA \\
        \cmidrule(lr){2-12}
        & EM & F1 & EM & F1 & EM & F1 & EM & F1 & EM & F1 & EM \\
        \midrule
        Naive RAG & \colorbox{MyLightBlue}{\textbf{38.20}} & 50.08 & \colorbox{MyLightBlue}{\textbf{60.80}} & 69.55 & 37.60 & 45.69 & 14.80 & 24.27 & 25.80 & 35.80 & 62.60 \\
        + $f_\text{RetDebate}$ & \colorbox{MyBlue}{\textbf{38.20}} & \colorbox{MyBlue}{\textbf{50.79}} & 57.20 & 65.31 & \colorbox{MyBlue}{\textbf{39.60}} & \colorbox{MyBlue}{\textbf{47.04}} & 16.00 & 25.91 & 28.20 & 37.48 & 64.60 \\
        + $f_\text{ResDebate}$ & 34.40 & 47.61 & 60.40 & \colorbox{MyLightBlue}{\textbf{69.60}} & 37.60 & 46.43 & \colorbox{MyLightBlue}{\textbf{26.00}} & \colorbox{MyLightBlue}{\textbf{34.30}} & \colorbox{MyLightBlue}{\textbf{30.20}} & \colorbox{MyLightBlue}{\textbf{41.00}} & \colorbox{MyLightBlue}{\textbf{68.80}} \\
        \midrule
        DRAG & 36.80 & \colorbox{MyLightBlue}{\textbf{50.38}} & \colorbox{MyBlue}{\textbf{60.80}} & \colorbox{MyBlue}{\textbf{69.93}} & \colorbox{MyLightBlue}{\textbf{38.60}} & \colorbox{MyLightBlue}{\textbf{46.50}} & \colorbox{MyBlue}{\textbf{28.80}} & \colorbox{MyBlue}{\textbf{36.97}} & \colorbox{MyBlue}{\textbf{30.80}} & \colorbox{MyBlue}{\textbf{41.74}} & \colorbox{MyBlue}{\textbf{69.20}} \\
        - \textit{Info Asymmetry} & 35.20 & 50.09 & 57.40 & 66.21 & 37.80 & 45.95 & 22.00 & 29.36 & 30.00 & 38.62 & 68.20 \\
        \bottomrule
    \end{tabular}
    }
    \caption{Ablation study for key components of DRAG on six benchmarks.  \textcolorblue{Blue} marks the \textcolorblue{best-performing} method, \textcolorlblue{light blue} represents the \textcolorlblue{second-best-performing} method.}
    \label{tab:ablation}
\end{table*}

\subsection{Ablation Study}

As shown in Table~\ref{tab:ablation}, we conduct an ablation study to evaluate the impact of key components in DRAG across six benchmarks.

First, we independently assess the effectiveness of the two debate stages: retrieval debate ($f_\text{RetDebate}$) and response debate ($f_\text{ResDebate}$).
To this end, we introduce two DRAG variants: (1) + $f_\text{RetDebate}$: Incorporates retrieval debate in the retrieval stage but excludes the Response Debate.
(2) + $f_\text{ResDebate}$: Incorporates response debate while omitting retrieval debate.
Compared to Naive RAG, both variants improve performance across most tasks. This confirms that the Retrieval Debate enhances retrieval strategies by reducing information bias and improving factual coverage, leading to better answer generation.
Meanwhile, response debate ensures proper utilization of retrieved information, mitigating hallucinations and enhancing factual accuracy.

To examine the effect of information asymmetry in response debate, we introduce the - \textit{Info Asymmetry} variant, where all agents have equal access to retrieval evidence.
Results show consistent performance deterioration across all datasets, indicating that information asymmetry is crucial for promoting factual consistency. 
By enforcing role-based knowledge distribution, it encourages adversarial interactions, prevents agents from over-relying on retrieved content, and improves robustness in factual reasoning.

\begin{figure}[t]
    \centering
    \includegraphics[width=0.98\linewidth]{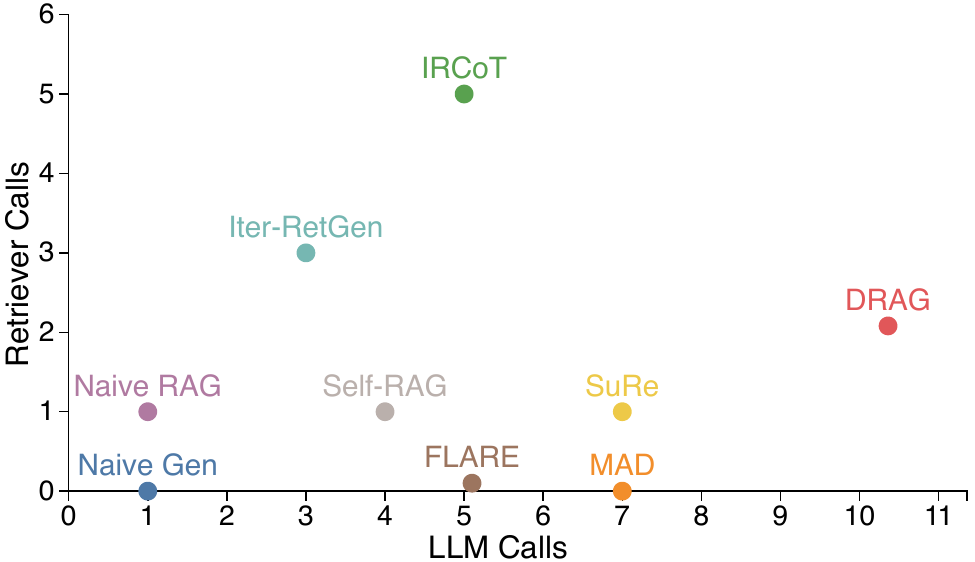}
    \caption{Average LLM and Retriever calls for DRAG and other baselines methods on the StrategyQA.}
    \label{fig:calls}
\end{figure}

\subsection{Efficiency Analysis}

We provide detailed comparisons of average LLM and retrieval calls on StrategyQA, as shown in Figure~\ref{fig:calls}. 
While DRAG requires more LLM calls than simpler baselines, making an average of 2.08 Retriever Calls and 10.36 LLM Calls (with 3.36 during the retrieval phase and 7 during the response phase), its overall cost remains comparable to other multi-turn frameworks such as IRCoT.
This cost is further mitigated by DRAG’s adaptive termination mechanism in the retrieval debate stage, which helps eliminate unnecessary debate rounds.
As a result, the number of LLM calls in the retrieval stage is significantly lower than in the response stage (7 rounds on average).
Moreover, as discussed in Section~\ref{sec:an_rounds}, for simpler single-hop questions, only 1–2 debate rounds are often sufficient.
In such cases, the number of LLM calls in the response stage can be reduced from 7 to 5 or fewer, achieving a more favorable trade-off between accuracy and efficiency.
A more effective adaptive stopping mechanism in the response phase may further enhance DRAG’s efficiency, and we will elaborate in the Limitations section.

\section{Conclusion}
In this paper, we introduce Debate-Augmented RAG (DRAG), a novel framework designed to mitigate hallucinations in Retrieval-Augmented Generation (RAG) by leveraging multi-agent debate mechanisms. 
DRAG employs a two-stage structured debate: (1) Retrieval Debate, where multiple agents iteratively refine retrieval queries to improve knowledge coverage;
(2) Response Debate, where agents adopt asymmetric roles and engage in structured debates to strengthen reasoning and mitigate reliance on flawed retrievals.
Extensive experiments across six datasets spanning three knowledge-intensive tasks validate the efficacy of DRAG.

\section*{Limitations}
A key limitation of our approach is the increased computational overhead introduced by multi-agent debates in both the retrieval and response stages.
While the retrieval phase employs a Judge Agent to dynamically terminate early when sufficient evidence is gathered, the response phase currently uses a fixed number of debate rounds, which may lead to unnecessary LLM calls, especially in simple single-hop tasks.

This also introduces another limitation: problem drift, where excessive reasoning adds unnecessary complexity and reduces effectiveness in straightforward scenarios. Although DRAG adapts retrieval depth well, its response stage lacks such flexibility.

Future work should explore adaptive stopping criteria, agent pruning, or confidence-based early termination in the response stage to better balance reasoning depth and efficiency. We believe these enhancements will improve DRAG's flexibility and performance across both single-hop and multi-hop settings. Investigating how to make the response debate more task-sensitive, by learning to infer when enough reasoning has been done, is a promising direction for future research.

\section*{Acknowledgement}
This work was supported by the National Natural Science Foundation of China (Grant No.: 62372314), Hong Kong Research Grants Council under General Research Fund (project no. 15200023), as well as Research Impact Fund (project no R1015-23).
This work was also supported by the following funding sources: industry donations from Accel Group Holding Limited and Minshang Creative Technology Holdings Limited (P0045948, P0046453); PolyU internal research funding (P0046701, P0046703); Research Matching Grant Scheme funded by the University Grants Committee (P0048183, P0048191); the Innovation and Technology Fund - ITSP, ITS/028/22FP (P0048887); Postdoc Matching Fund Scheme (P0048984); the RGC Early Career Scheme, 25600624 (P0051906); and Two Square Capital Limited donation (P0054482).

\bibliography{custom}

\appendix
\clearpage
\section{Experimental Details}

\subsection{More Implementation Details}\label{appendix_implementation_details}
We run our models on eight NVIDIA RTX A6000 GPU devices. To get a better performance, for IRCoT, we set the iteration number to 5. 
For all baselines, we reproduced the key settings from their original papers as faithfully as possible under partially unified conditions, including the few-shot setting. 
In contrast, our DRAG performs inference under a zero-shot setting.
As a training-based approach, Self-RAG utilizes the Self-RAG-Llama-2-7B\footnote{\url{https://huggingface.co/selfrag/selfrag_llama2_7b}} model provided by its authors for experimentation.
Additionally, since SuRe and Self-RAG are not well-suited for answering questions with definitive yes/no answers, their performance is relatively poor.

\subsection{Prompts for Retrieval Debate}

An example of a proponent agent prompt in the Retrieval Debate stage is 
\begin{tcolorbox}[colback=black!1!white,colframe=black!57!white,title=Prompt for Proponent Agent, breakable]
You are a debater. Argue that the current retrieved content is sufficient to answer the question and no further retrieval is needed. Deliver a brief, strong argument with clear reasoning. Do not suggest further retrieval.\\
\\
Question:\\
\{question\}\\
\\
Queries:\\
\{queries in $Q$\}\\
\\
Retrieved Documents:\\
\{retrieved results in $C_Q$\}\\
\end{tcolorbox}

In the Retrieval Debate stage, the challenger agent is responsible for proposing modifications to unreasonable queries or initiating new retrieval queries to address further knowledge needs. Inspired by Toolformer \cite{schick2023toolformer} and FLARE \cite{jiang-etal-2023-active}, we can directly prompt the agent to choose two types of operations: {\tt Query Optimization} and {\tt Query Expansion}, and an example prompt is 
\begin{tcolorbox}[colback=black!1!white,colframe=black!57!white,title=Prompt for Challenger Agent, breakable]
You are a critical thinker and debater, and your task is to challenge the sufficiency of the current retrieved content. Argue that the current information is insufficient to generate a reliable answer and propose either query optimization or query expansion.\\
The action you can choose:\\
1. {\tt Query Optimization}: If the retrieved content is somewhat relevant but has expression or scope issues.\\
Optimize the query using this format: Query Optimization: [Original Query] $\rightarrow$ [New Query].\\
2. {\tt Query Expansion}: If critical information is missing.\\
Propose a new query using this format: Query Expansion: [New Query].\\
\\
Deliver a brief, strong argument with clear reasoning, and then you must choose only one action. The output must be in the exact format after your reasoning, without additional explanation, and keep the new query short and precise.\\
\\
Question:\\
\{question\}\\
\\
Queries:\\
\{queries in $Q$\}\\
\\
Retrieved Documents:\\
\{retrieved results in $C_Q$\}\\
\end{tcolorbox}

Instead of calculating the convergence conditions, we directly use the judge agent to evaluate the arguments presented by the proponent agent and the challenger agent and determine the winning agent for the current round of debate.
An example prompt is
\begin{tcolorbox}[colback=black!1!white,colframe=black!57!white,title=Prompt for Judge Agent, breakable]
You are the judge in a debate. Your task is to evaluate the arguments from agents.\\
There are two types of agents:\\
1. Proponent Agent: Argue that the current retrieved content is sufficient.\\
2. Challenger Agent: Argue that the current retrieved content is insufficient and propose query refinement.\\
\\
Question:\\
\{question\}\\
\\
Queries:\\
\{queries in $Q$\}\\
\\
Retrieved Documents:\\
\{retrieved results in $C_Q$\}\\
\\
Agents Arguments:\\
\{agents arguments\}\\
\\
Output only the agent's name.
\end{tcolorbox}

\subsection{Prompts for Response Debate}
In the Response Debate phase, the discussion is initialized based on the retrieval pool and its associated documents produced in the Retrieval Debate phase.
An example of a proponent agent prompt in the Response Debate stage is
\begin{tcolorbox}[colback=black!1!white,colframe=black!57!white,title=Prompt for Initializing Proponent Agent, breakable]
Answer the question based on the given document. \\

The following are given documents:\\
Query $1$: $q_1 \in Q$\\
Retrieved Documents: $C_{q_1}\in C_Q$\\
\\
Query $2$: $q_2 \in Q$\\
Retrieved Documents: $C_{q_2}\in C_Q$\\
...\\
Query $n$: $q_n \in Q$\\
Retrieved Documents: $C_{q_n}\in C_Q$\\

Question:\\
\{question\}
\end{tcolorbox}
The challenger agent is initialized based on internal knowledge, and an example of a challenger agent prompt is
\begin{tcolorbox}[colback=black!1!white,colframe=black!57!white,title=Prompt for Initializing Challenger Agent, breakable]
Answer the question based on your own knowledge. \\

Question:\\
\{question\}
\end{tcolorbox}
Factual consistency in the Response Debate phase is achieved through iterative exchanges and debates among agents. An example of the debate prompt is
\begin{tcolorbox}[colback=black!1!white,colframe=black!57!white,title=Prompt for Debate, breakable]
I will give the answers and arguments to this question from other agents. Use their solution as additional advice; note that they may be wrong.\\
Explain your answer.\\

\{Agent's Name\}:\\
\{Other agent's response\}\\

Question:\\
\{question\}
\end{tcolorbox}
In the Response Debate phase, the Judge Agent aggregates the responses from all agents and selects the most probable answer as the final output. An example of a judge agent prompt is
\begin{tcolorbox}[colback=black!1!white,colframe=black!57!white,title=Prompt for Judge Agent, breakable]
You are a moderator in a debate competition. Your task is to determine the correct final answer based on the arguments presented by the debaters. Output only the final answer with no explanations or additional text.\\

\{Agent's Name\}:\\
\{Agent's Response\}\\

Question:\\
\{question\}

\end{tcolorbox}

\begin{table*}[t]
    \centering
    \adjustbox{max width=\textwidth}{
     \begin{tabular}{lcc|cc|cc|cc|cc|c}
        \toprule
        \multirow{2}{*}{Settings} & \multicolumn{2}{c}{NQ} & \multicolumn{2}{c}{TriviaQA} & \multicolumn{2}{c}{PopQA} & \multicolumn{2}{c}{2Wiki} & \multicolumn{2}{c}{HotpotQA} & StrategyQA \\
        \cmidrule(lr){2-12}
        & EM & F1 & EM & F1 & EM & F1 & EM & F1 & EM & F1 & EM \\
        \midrule
        DRAG$_{\text{RetIter=0}}$& 34.40 & 47.61 & 60.40 & 69.60 & 37.60 & 45.43 & 26.00 & 34.30 & 30.20 & 41.00 & 68.80 \\
        DRAG$_{\text{RetIter=1}}$& 36.80 & \textbf{50.64} & 60.80 & 69.90 & 37.40 & 45.60 & 29.20 & 37.49 & \textbf{31.40} & \textbf{42.41} & \textbf{69.60} \\
        DRAG$_{\text{RetIter=2}}$& 36.80 & 50.32 & 60.76 & 69.93 & 37.50 & 45.83 & \textbf{29.60} & \textbf{37.84} & 30.80 & 41.85 & 69.40 \\
        DRAG$_{\text{RetIter=3}}$& 36.80 & 50.38 & \textbf{60.80} & \textbf{69.93} & \textbf{37.60} & \textbf{46.10} & 28.80 & 36.97 & 30.80 & 41.74 & 69.20 \\
        DRAG$_{\text{RetIter=4}}$& \textbf{37.00} & 50.58 & 60.80 & 69.91 & 37.60 & 46.10 & 29.00 & 37.33 & 31.00 & 41.86 & 69.40 \\
        \midrule
        DRAG$_{\text{ResIter=0}}$& 38.20 & 50.79 & 57.20 & 65.31 & \textbf{39.60} & 47.04 & 16.00 & 25.91 & 28.20 & 37.48 & 64.60 \\
        DRAG$_{\text{ResIter=1}}$& 37.60 & \textbf{51.77} & \textbf{62.40} & \textbf{71.47} & 39.20 & \textbf{47.52} & 27.80 & 36.91 & \textbf{33.40} & \textbf{43.19} & \textbf{71.00} \\
        DRAG$_{\text{ResIter=2}}$& \textbf{38.20} & 51.61 & 60.20 & 68.59 & 38.80 & 46.52 & 28.20 & 36.47 & 31.20 & 41.53 & 68.00 \\
        DRAG$_{\text{ResIter=3}}$& 36.80 & 50.38 & 60.80 & 69.93 & 38.60 & 46.50 & \textbf{28.80} & \textbf{36.97} & 30.80 & 41.74 & 69.20 \\
        DRAG$_{\text{ResIter=4}}$& 37.00 & 50.09 & 60.60 & 70.03 & 38.80 & 46.98 & 27.20 & 34.67 & 31.00 & 41.30 & 67.00 \\
        \bottomrule
    \end{tabular}
    }
    \caption{The detailed performance metrics of DRAG across six datasets under different debate round settings in the Retrieval Debate and Response Debate stages.}
    \label{tab:eval_rounds_detail}
\end{table*}

\section{More Detailed Results}
Table~\ref{tab:eval_rounds_detail} shows the detailed performance metrics of DRAG across six datasets under different debate round settings in the Retrieval Debate and Response Debate stages.

\begin{figure*}[t]
    \centering
    \includegraphics[width=0.9\linewidth]{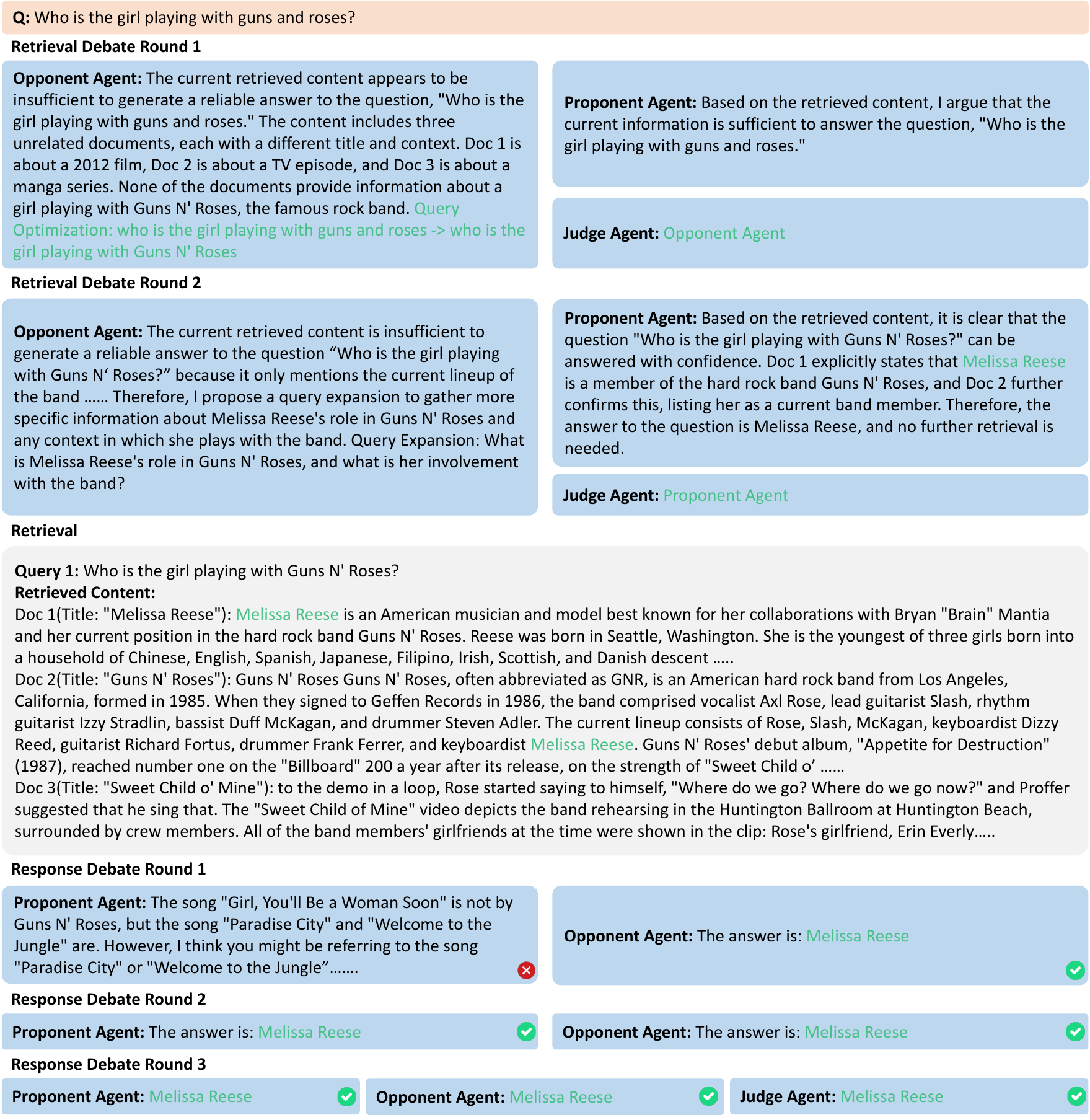}
    \caption{Case study of the Retrieval Debate. }
    \label{fig:case_study2}
\end{figure*}

\begin{figure*}[t]
    \centering
    \includegraphics[width=0.9\linewidth]{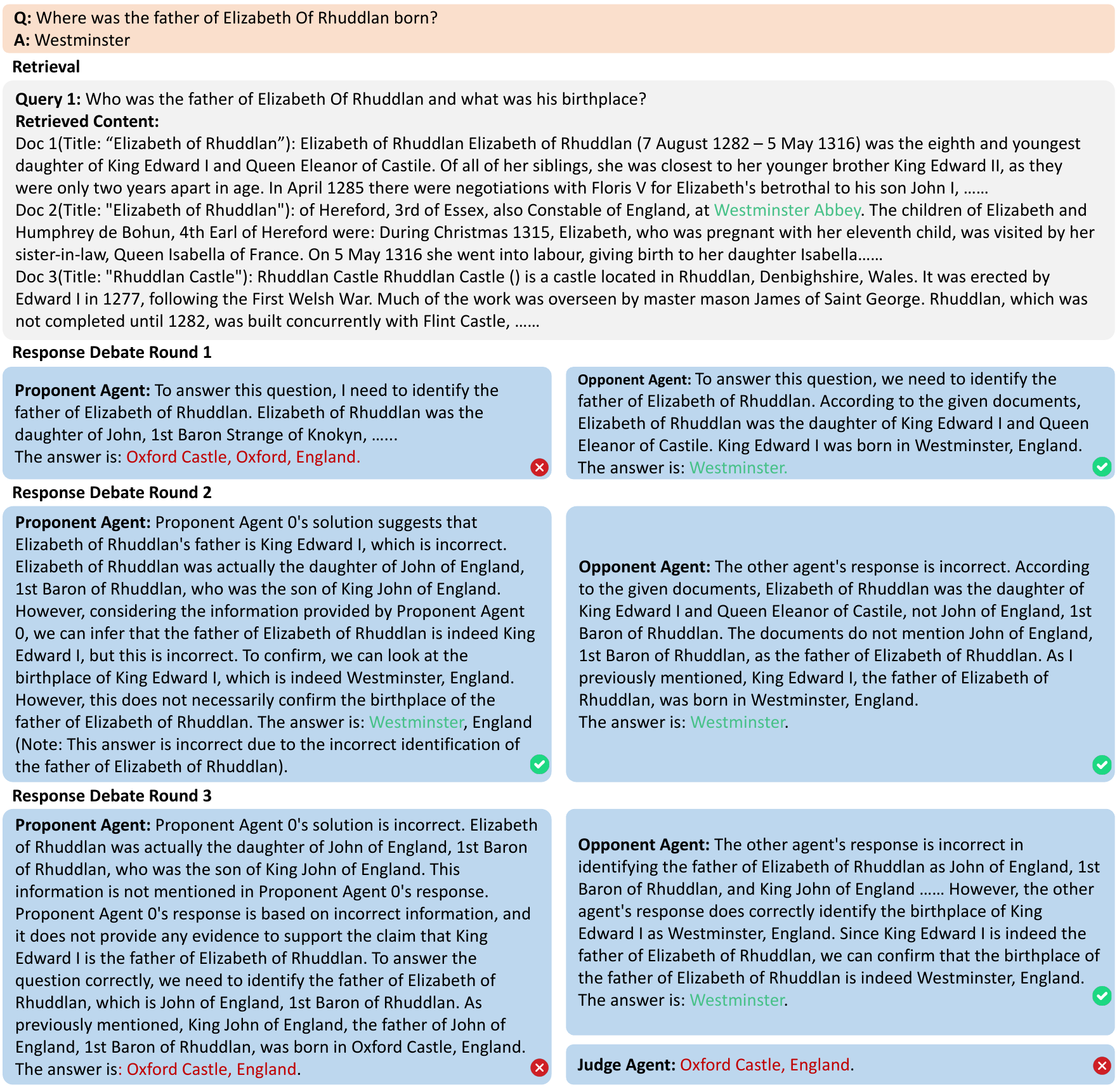}
    \caption{Error Case of the Response Debate. }
    \label{fig:case_study3}
\end{figure*}

\section{Case Study}
We show the case study of the Retrieval Debate in Figure~\ref{fig:case_study2}, and the error case in the Response Debate stage in Figure~\ref{fig:case_study3}. 

\end{document}